\documentclass[conference]{IEEEtran}
\IEEEoverridecommandlockouts
\usepackage{cite}
\usepackage{amsmath,amssymb,amsfonts}
\usepackage{algorithmic}
\usepackage{graphicx}
\usepackage{textcomp}
\usepackage{xcolor}
\usepackage[nottoc]{tocbibind}
\usepackage[ruled,longend,linesnumbered,lined,commentsnumbered]{algorithm2e}
\usepackage{soul}
\usepackage{color}
\usepackage{subfig}
\usepackage[export]{adjustbox}
\usepackage{float}
\usepackage{booktabs}
\usepackage{array}
\usepackage{siunitx}
\usepackage{nccmath}
\usepackage{caption}
\usepackage[most]{tcolorbox}
\usepackage{multirow}
\usepackage[numbers,sort&compress]{natbib}
\usepackage{mathtools}
\usepackage{scalerel}

\newcommand{\colvec}[2][.8]{%
  \scalebox{#1}{%
    \renewcommand{\arraystretch}{.8}%
    $\begin{bmatrix}#2\end{bmatrix}$%
  }
}

\newcolumntype{P}[1]{>{\centering\arraybackslash}p{#1}}

\def\BibTeX{{\rm B\kern-.05em{\sc i\kern-.025em b}\kern-.08em
    T\kern-.1667em\lower.7ex\hbox{E}\kern-.125emX}}
\begin{document}

\title{Object Detection and Pose Estimation from RGB and Depth Data for Real-time, Adaptive Robotic Grasping\\
}

\author{Shuvo Kumar Paul$^{1}$, Muhammed Tawfiq Chowdhury$^{1}$, Mircea Nicolescu$^{1}$, Monica Nicolescu$^{1}$, \\David Feil-Seifer$^{1}$ 

\thanks{*This work has been supported in part by the Office of Naval Research award N00014-16-1-2312 and US Army Research Laboratory (ARO) award W911NF-20-2-0084.}
\thanks{$^{1}$\textbf{Contact author: Shuvo Kumar Paul}, Muhammed Tawfiq Chowdhury, Mircea Nicolescu, Monica Nicolescu, and David Feil-Seifer are affiliated with the Department of Computer Science and Engineering,
        University of Nevada, Reno, 1664 North Virginia Street, Reno, Nevada 89557, USA
        {\tt\small  shuvo.k.paul@nevada.unr.edu,
        mtawfiqc@nevada.unr.edu, mircea@cse.unr.edu, monica@cse.unr.edu,
        dave@cse.unr.edu}}%
}
\maketitle

\begin{abstract}
In recent times, object detection and pose estimation have gained significant attention in the context of robotic vision applications. Both the identification of objects of interest as well as the estimation of their pose remain important capabilities in order for robots to provide effective assistance for numerous robotic applications ranging from household tasks to industrial manipulation. This problem is particularly challenging because of the heterogeneity of objects having different and potentially complex shapes, and the difficulties arising due to background clutter and partial occlusions between objects. As the main contribution of this work, we propose a system that performs real-time object detection and pose estimation, for the purpose of dynamic robot grasping. The robot has been pre-trained to perform a small set of canonical grasps from a few fixed poses for each object. When presented with an unknown object in an arbitrary pose, the proposed approach allows the robot to detect the object identity and its actual pose, and then adapt a canonical grasp in order to be used with the new pose. For training, the system defines a canonical grasp by capturing the relative pose of an object with respect to the gripper attached to the robot’s wrist. During testing, once a new pose is detected, a canonical grasp for the object is identified and then dynamically adapted by adjusting the robot arm's joint angles, so that the gripper can grasp the object in its new pose. We conducted experiments using a humanoid PR2 robot and showed that the proposed framework can detect well-textured objects, and provide accurate pose estimation in the presence of tolerable amounts of out-of-plane rotation. The performance is also illustrated by the robot successfully grasping objects from a wide range of arbitrary poses.

\end{abstract}
\begin{IEEEkeywords}
pose estimation, robotics, robotic grasp, homography
\end{IEEEkeywords}

\section{Introduction}
Current advances in robotics and autonomous systems have expanded the use of robots in a wide range of robotic tasks including assembly, advanced manufacturing, human-robot or robot-robot collaboration. In order for robots to efficiently perform these tasks, they need to have the ability to adapt to the changing environment while interacting with their surroundings, and a key component of this interaction is the reliable grasping of arbitrary objects. Consequently, a recent trend in robotics research has focused on object detection and pose estimation for the purpose of dynamic robotic grasping.

However, identifying objects and recovering their poses are particularly challenging tasks as objects in the real world are extremely varied in shape and appearance. Moreover, cluttered scenes, occlusion between objects, and variance in lighting conditions make it even more difficult. Additionally, the system needs to be sufficiently fast to facilitate real-time robotic tasks. As a result, a generic solution that can address all these problems remains an open challenge.

While classification~\cite{resnet, vgg, inception,flexnet, overfeat, spatial}, detection~\cite{fast-rcnn, faster-rcnn, ssd, yolo, yolo-9000, fldnet}, and segmentation~\cite{segnet, maskrcnn, unet}  of objects from images have taken a significant step forward - thanks to deep learning, the same has not yet happened to 3D localization and pose estimation. One primary reason was the lack of labeled data in the past as it is not practical to manually infer, thus   As a result, the recent research trend in the deep learning community for such applications has shifted towards synthetic datasets~\cite{butler,mayer,qiu,zhang,mccormac}. Several pose estimation methods leveraging deep learning techniques~\cite{posecnn,dope,brachmann,wang,hu} use these synthetic datasets for training and have shown satisfactory accuracy. 

Although synthetic data is a promising alternative, capable of generating large amounts of labeled data, it requires photorealistic 3D models of the objects to mirror the real-world scenario. Hence, generating synthetic data for each newly introduced object needs photo-realistic 3D models and thus significant effort from skilled 3D artists. Furthermore, training and running deep learning models are not feasible without high computing resources as well. As a result, object detection and pose estimation in real-time with computationally moderate machines remain a challenging problem. To address these issues, we have devised a simpler pipeline that does not rely on high computing resources and focuses on planar objects, requiring only an RGB image and the depth information in order to infer real-time object detection and pose estimation.

In this work, we present a feature-detector-descriptor based method for detection and a homography based pose estimation technique where, by utilizing the depth information, we estimate the pose of an object in terms of a 2D planar representation in 3D space. The robot is pre-trained to perform a set of canonical grasps; a canonical grasp describes how a robotic end-effector should be placed relative to an object in a fixed pose so that it can securely grasp it. Afterward, the robot is able to detect objects and estimates their pose in real-time, and then adapt the pre-trained canonical grasp to the new pose of the object of interest. We demonstrate that the proposed method can detect a well-textured planar object and estimate its accurate pose within a tolerable amount of out-of-plane rotation. We also conducted experiments with the humanoid PR2 robot to show the applicability of the framework where the robot grasped objects by adapting to a range of different poses.


\section{Related Work}
Our work constitutes of three modules: object detection, planar pose estimation, and adaptive grasping. In the following sub-sections, several fields of research that are closely related to our work are reviewed.
\subsection{Object Detection}
Object detection has been one of the fundamental challenges in the field of computer vision and in that aspect, the introduction of feature detectors and descriptors represents a great achievement. Over the past decades, many detectors, descriptors, and their numerous variants have been presented in the literature. The applications of these methods have widely extended to numerous other vision applications such as panorama stitching, tracking, visual navigation, etc.

One of the first feature detectors was proposed by Harris et al.~\cite{harris} (widely known as the Harris corner detector). Later Tomasi et al.~\cite{tomasi} developed the KLT (Kanade-Lucas-Tomasi) tracker based on the Harris corner detector. Shi and Tomasi introduced a new detection metric GFTT~\cite{shi} (Good Features To Track) and argued that it offered superior performance. Hall et al. introduced the concept of saliency~\cite{hall} in terms of the change in scale and evaluated the Harris method proposed in~\cite{lindeberg} and the Harris Laplacian corner detector~\cite{mikolajczyk} where a Harris detector and a Laplacian function are combined.

Motivated by the need for a scale-invariant feature detector, in 2004 Lowe~\cite{SIFT} published one of the most influential papers in computer vision, SIFT (Scale Invariant Feature Transform). SIFT is both a feature point detector and descriptor. H. Bay et al.~\cite{SURF} proposed SURF (Speeded Up Robust Features) in 2008. But both of these methods are computationally expensive as SIFT detector leverages the difference of Gaussians (DoG) in different scales while SURF detector uses a Haar wavelet approximation of the determinant of the Hessian matrix to speed up the detection process. Many variants of SIFT~\cite{sift_v_1, sift_v_2, sift_v_3, sift_v_4} and SURF~\cite{surf_v_1, surf_v_2, surf_v_3} were proposed, either targeting a different problem or reporting improvements in matching, however, the execution time remained a persisting problem for several vision applications.

To improve execution time, several other detectors such as FAST~\cite{fast} and AGAST~\cite{agast} have been introduced. Calonder et al. developed the BRIEF~\cite{brief} (Binary Robust Independent Elementary Features) descriptor of binary strings that has a fast execution time and is very useful for matching images. E. Rublee et al. presented ORB~\cite{ORB} (Oriented FAST and Rotated Brief) which is a combination of modified FAST (Features from Accelerated Segment Test) for feature detection and BRIEF for description. S. Leutnegger et al. designed BRISK~\cite{BRISK} (Binary Robust Invariant Scale Keypoint) that detects corners using AGAST and filters them using FAST. On the other hand, FREAK (Fast Retina Key-point), introduced by Alahi et al.~\cite{FREAK} generates retinal sampling patterns using a circular sampling grid and uses a binary descriptor, formed by a one bit difference of Gaussians (DoG). Alcantarilla et al. introduced KAZE~\cite{KAZE} features that exploit non-linear scale-space using non-linear diffusion filtering and later extended it to AKAZE~\cite{AKAZE} where they replaced it with a more computationally efficient method called FED (Fast Explicit Diffusion)~\cite{fed1, fed2}.

In our work, we have selected four methods to investigate: SIFT, SURF, FAST+BRISK, AKAZE. 

\subsection{Planar Pose Estimation}

Among the many techniques in literature on pose estimation, we focus our review on those related to planar pose estimation. In recent years, planar pose estimation has been increasingly becoming popular in many fields, such as robotics and augmented reality. 

Simon et. al~\cite{simon} proposed a pose estimation technique for planar structures using homography projection and by computing camera pose from consecutive images. Changhai et. al~\cite{changhai} presented a method to robustly estimate 3D poses of planes by applying a weighted incremental normal estimation method that uses Bayesian inference. Donoser et al.~\cite{donoser} utilized the properties of Maximally Stable Extremal Regions (MSERs~\cite{LMSER}) to construct a perspectively invariant frame on the closed contour to estimate the planar pose. In our approach, we applied perspective transformation to approximate a set of corresponding points on the test image for estimating the basis vectors of the object surface and used the depth information to estimate the 3D pose by computing the normal to the planar object.

\subsection{Adaptive Grasping}

Designing an adaptive grasping system is challenging due to the complex nature of the shapes of objects. In early times, analytical methods were used where the system would analyze the geometric structure of the object and would try to predict suitable grasping points. Sahbani et al.~\cite{sahbani} did an in depth review on the existing analytical approaches for 3D object grasping. However, with the analytical approach it is difficult to compute force and not suitable for autonomous manipulation. Later, as the number of 3D models increased, numerous data driven methods were introduced that would analyze grasps in the 3D model database and then transfer to the target object. Bohg et al.~\cite{bohg} reviewed data driven grasping method methods where they divided the approach into three groups based on the familiarity of the object.

Kehoe et al. \cite{Kehoe2013} used a candidate grasp from the candidate grasp set based on the feasibility score determined by the grasp planner. The grasps weren't very accurate in situations where the objects had stable horizontal poses and were close to the width of the robot's gripper. Huebner et al. \cite{Huebner2008} also take a similar approach as they perform grasp candidate simulation. They created a sequence of grasps by approximating the shape of the objects and then computed a random grasp evaluation for each model of objects. In both works, a grasp has been chosen from a list of candidate grasps.

The recent advances in deep learning also made it possible to regress grasp configuration through deep convolutional networks. A number of deep learning-based methods were reviewed in~\cite{caldera} where the authors also discussed how each element in deep learning-based methods enhances the robotic grasping detection. \cite{Yu2013} presented a system where deep neural networks were used to learn hierarchical features to detect and estimate the pose of an object, and then use the centers of the defined pose classes to grasps the objects. Kroemer et al.~\cite{Kroemer2009} introduced an active learning approach where the robot observes a few good grasps by demonstration and learns a value function for these grasps using Gaussian process regression. Aleotti et al.~\cite{Aleotti2011} proposed a grasping model that is capable of grasping objects by their parts which learns new tasks from human demonstration with automatic 3D shape segmentation for object recognition and semantic modeling. \cite{Saxena2008} and \cite{Montesano2012} used supervised learning to predict grasp locations from RGB images. In~\cite{Nogueira2016}, as an alternative to a trial and error exploration strategy, the authors proposed a Bayesian optimization technique to address the robot grasp optimization problem of unknown objects. These methods emphasized developing and using learning models for obtaining accurate grasps. 

In our work, we focus on pre-defining a suitable grasp relative to an object that can adapt to a new grasp based on the change of position and orientation of the object.

\section{Method}
\begin{figure*}
\begin{center}
\includegraphics[width=0.85\linewidth, height=0.50\linewidth]{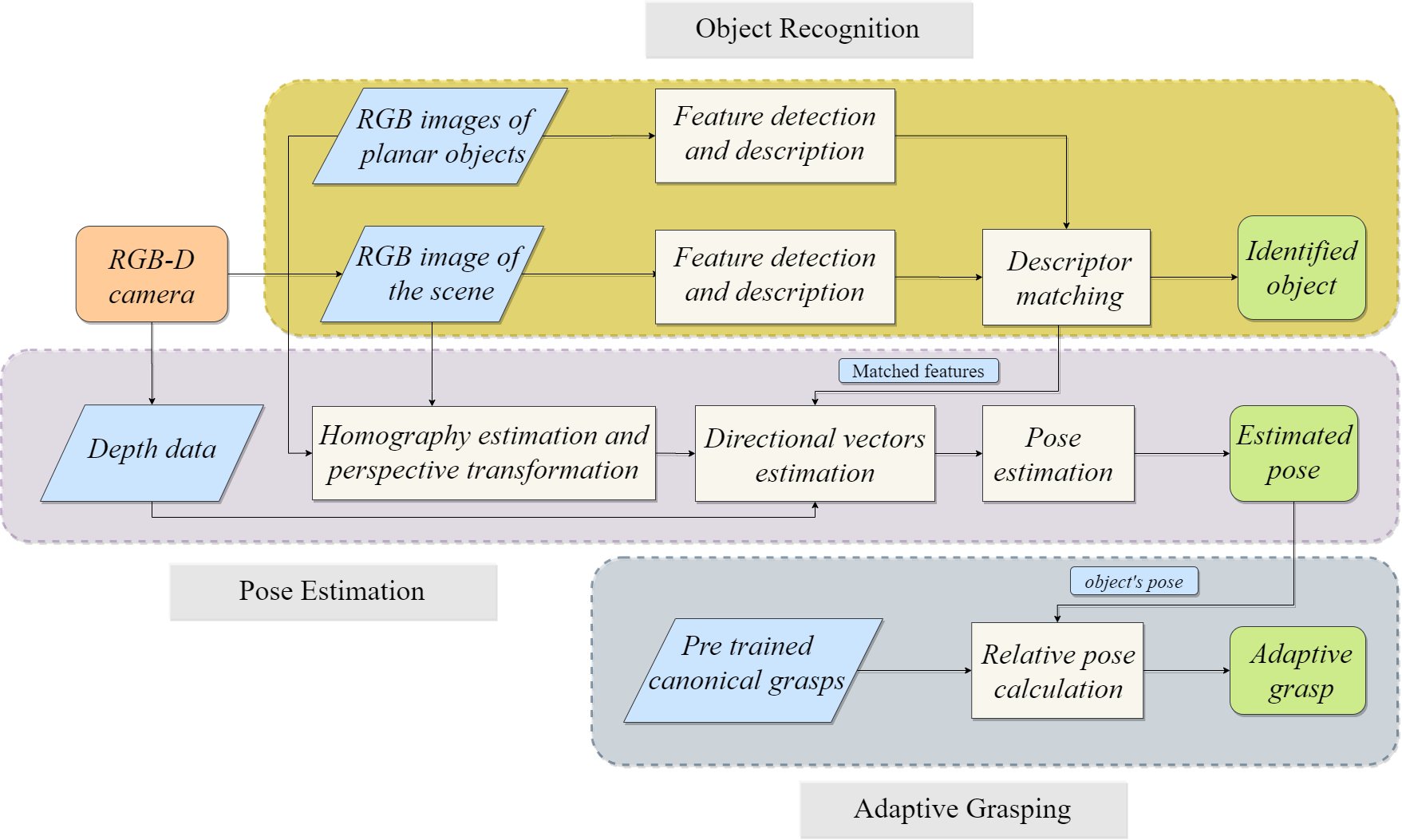}
\end{center}
  \caption{System architecture.}
\label{fig:sysarch}

\end{figure*}

The proposed method is divided into two parts. The first part outlines the process of simultaneous object detection and pose estimation of multiple objects and the second part describes the process of generating an adaptive grasp using the pre-trained canonical grasp and the object pose.
The following sections describe the architecture of the proposed framework (figure~\ref{fig:sysarch}) in detail.

\subsection{Object Detection and Pose Estimation}

We present a planar pose estimation algorithm (algorithm \ref{algorithm}) for adaptive grasping that consists of four phases: (i) feature extraction and matching, (ii) homography estimation and perspective transformation, (iii) directional vectors estimation on the object surface, (iv) planar pose estimation using the depth data. In the following sections, we will focus on the detailed description of the aforementioned steps.

\subsubsection{Feature extraction and matching}

Our object detection starts with extracting features from the images of the planar objects and then matching them with the features found in the images acquired from the camera. Image features are patterns in images based on which we can describe the image. A feature detecting algorithm takes an image and returns the locations of these patterns - they can be edges, corners or interest points, blobs or regions of interest points, ridges, etc. This feature information then needs to be transformed into a vector space using a feature descriptor, so that it gives us the possibility to execute numerical operations on them. A feature descriptor encodes these patterns into a series of numerical values that can be used to match, compare, and differentiate one feature to another; for example, we can use these feature vectors to find the similarities in different images which can lead us to detect objects in the image. In theory, this information would be invariant to image transformations. In our work, we have investigated SIFT~\cite{SIFT}, SURF~\cite{SURF}, AKAZE~\cite{AKAZE}, and BRISK~\cite{BRISK} descriptors. SIFT, SURF, AKAZE are both feature detectors and descriptors, but BRISK uses FAST~\cite{fast} algorithm for feature detection. These descriptors were selected after carefully reviewing the comparisons done in the recent literature~\cite{andersson2016comparison, karami2017image, tareen2018comparative}.

Once the features are extracted and transformed into vectors, we compare the features to determine the presence of an object in the scene. For non-binary feature descriptors (SIFT, SURF) we find matches using the Nearest Neighbor algorithm. However, finding the nearest neighbor matches within high dimensional data is computationally expensive, and with more objects introduced it can affect the process of updating the pose in real-time. To counter this issue to some extent, we used the FLANN~\cite{muja_flann_2009} implementation of K-d Nearest Neighbor Search, which is an approximation of the K-Nearest Neighbor algorithm that is optimized for high dimensional features. For binary features (AKAZE, BRISK), we used the Hamming distance ratio method to find the matches. Finally, if we have more than ten matches, we presume the object is present in the scene. 

\RestyleAlgo{boxruled}
\begin{algorithm}[ht]
\fontsize{8}{8}\selectfont
\DontPrintSemicolon
    \KwIn{Training images of planar objects, $\mathcal{I}$}
    $Detector \gets \text{Define feature detector}$\;
    $Descriptor \gets \text{Define feature descriptor}$\;
    \tcc{\fontsize{8}{8}\selectfont retrieve feature descriptor}
    \tcc{\fontsize{8}{8}\selectfont for each image in $\mathcal{I}$}
    \For{i in $\mathcal{I}$}{
        \tcc{\fontsize{7}{7}\selectfont $\mathcal{K}$ is set of detected keypoints for image i}
       \fontsize{8}{8}\selectfont $\mathcal{K} \gets \texttt{DetectKeypoints($i, Detector$)}$\;
        \tcc{\fontsize{7}{7}\selectfont $\mathcal{D}[i]$ is the corresponding descriptor set for image i }
        $\mathcal{D}[i] \gets \texttt{GetDescriptors( $\mathcal{K}, Descriptor$)}$\;
    }

    \While{$\text{camera is on}$}
    { \fontsize{8}{8}\selectfont
        $f \gets \text{RGB image frame}$\;
        $PC \gets \text{Point cloud data}$\;
        \tcc{\fontsize{7}{7}\selectfont $K_F$ is set of detected keypoints for image frame $f$}
        $K_F \gets \texttt{DetectKeypoints($f, Detector$)}$\;
        \tcc{\fontsize{7}{7}\selectfont $D_F$ is the corresponding descriptor set for rgb image $f$}
        $D_F \gets \texttt{GetDescriptors( $K_F, Descriptor$)}$\;
        \For{i in $\mathcal{I}$}
        {
            $matches \gets \texttt{FindMatches( $\mathcal{D}[i]$, $D_F$)}$\;
            \tcc{\fontsize{7}{7}\selectfont If there is at least 10 matches then we have the object (described in image $i$) in the scene}
            \uIf{\text{Total number of }$matches \geq 10$}
            {
                \tcc{\fontsize{7}{7}\selectfont extract matched keypoints pair $(kp_{i},kp_{f})$ from the corresponding descriptors matches.}
                $kp_{i}, kp_{f} \gets \texttt{ExtractKeypoints($matches$)}$\;
                $\mathbf{H} \gets \texttt{EstimateHomography($kp_{i}, kp_{f}$)}$\;
                $p_c, p_x, p_y \gets \text{points on the planar object }\newline \text{~~~~~~~~~~~~~ obtained using  equation (\ref{eqn:axis})}$\;
                $p_c^{'}, p_x^{'}, p_y^{'} \gets \text{corresponding projected points}\newline \text{~~~~~~~~~~~~~ of $p_c, p_x, p_y$ on image frame $f$}\newline \text{~~~~~~~~~~~~~  estimated using equations}\newline \text{~~~~~~~~~~~~~ (\ref{eqn:homography}) and (\ref{eqn:projection})}$\;
                \tcc{\fontsize{7}{7}\selectfont $\vec{c}$ denotes the origin of the object frame with respect to the base/world frame}
                $\Vec{c}, \Vec{x}, \Vec{y} \gets \text{corresponding 3d locations }\newline \text{~~~~~~~~~ of $p_c^{'}, p_x^{'}, p_y^{'}$ from point cloud $PC$}$\;
                \tcc{\fontsize{7}{7}\selectfont shift $\vec{x}, \vec{y}$ to the  origin of the base or the world frame}
                $\vec{x} \gets \vec{x}-\vec{c}$\; 
                $\vec{y} \gets \vec{y}-\vec{c}$\;
                \tcc{\fontsize{7}{7}\selectfont estimate the object frame in terms of three orthonormal vectors 
                $\hat{i}, \hat{j}$, and $\hat{k}$.}
                $\hat{i}, \hat{j}, \hat{k} \gets \text{from equation (\ref{eqn:unitv})}$\;
                \tcc{\fontsize{7}{7}\selectfont compute the rotation  $\phi_i,\theta_i,\psi_i$ of the object frame $\hat{i}, \hat{j}$, $\hat{k}$ with respect to the base or the world frame $\vec{X}, \vec{Y}, \vec{Z}$.}
                $\phi_i,\theta_i,\psi_i \gets \text{from equation (\ref{eqn:eulerangles})}$\;
                \tcc{\fontsize{7}{7}\selectfont finally, publish the position and orientation of the object.}
                \texttt{publish$(\vec{c},\phi_i,\theta_i,\psi_i)$}\;
            }
        }
    }
  \caption{Planar Pose Estimation}
  \label{algorithm}
\end{algorithm}

\subsubsection{Homography Estimation and Perspective Transformation}
\begin{figure}[h]
\begin{center}
\graphicspath{ {./images/} }
\includegraphics[height=6cm]{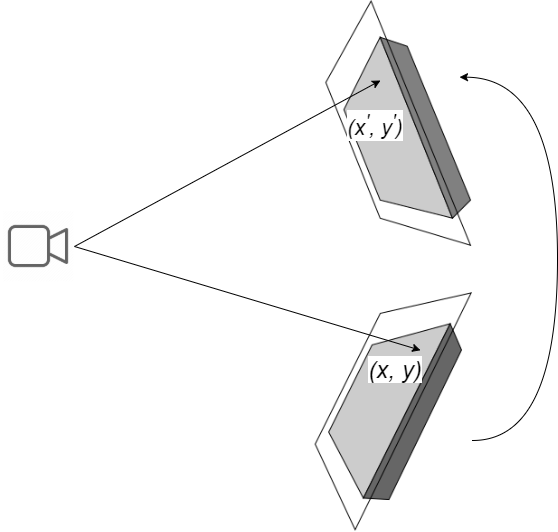}
\end{center}
  \caption{Object in different orientation from the camera}
\label{fig:homography}
\end{figure}

A homography is an invertible mapping of points and lines on the projective plane that describes a 2D planar projective transformation~(figure~\ref{fig:homography}) that can be estimated from a given pair of images. In simple terms, a homography is a matrix that maps a set of points in one image to the corresponding set of points in another image. We can use a homography matrix $\mathbf{H}$ to find the corresponding points using equation \ref{eqn:homography} and~\ref{eqn:projection}, which defines the relation of projected point $(x^{'}, y^{'})$ (figure \ref{fig:homography}) on the rotated plane to the reference point $(x,y)$. 

A 2D point $(x,y)$ in an image can be represented as a 3D vector $(x, y, 1)$ which is called the homogeneous representation of a point that lies on the reference plane or image of the planar object. In equation (\ref{eqn:homography}), $\mathbf{H}$ represents the homography matrix and $[x~y~1]^{T}$ is the homogeneous representation of the reference point $(x,y)$ and we can use the values of $a,b,c$ to estimate the projected point $(x^{'},y^{'})$ in equation (\ref{eqn:projection}). 

\begin{align}
 \left [ \begin{matrix} a \\ b \\  c \end{matrix} \right ] = \mathbf{H}\begin{bmatrix} x\\ y\\ 1\\ \end{bmatrix} = \begin{bmatrix} h_{11}&h_{12}&h_{13}\\ h_{21}&h_{22}&h_{23}\\ h_{31}&h_{32}&h_{33}\\ \end{bmatrix} \begin{bmatrix} x\\ y\\ 1\\ \end{bmatrix}
\label{eqn:homography}   
\end{align}

\begin{equation}
    \begin{aligned}
\left \lbrace \begin{aligned}
    x^{'} = \frac{a}{c} \\
    y^{'} = \frac{b}{c}  
\end{aligned} \right . 
\end{aligned}
\label{eqn:projection}
\end{equation}

We estimate the homography using the matches found from the nearest neighbor search as input; often these matches can have completely false correspondences, meaning they don't correspond to the same real-world feature at all which can be a problem in estimating the homography. So, we chose RANSAC~\cite{ransac} to robustly estimate the homography by considering only inlier matches as it tries to estimate the underlying model parameters and detect outliers by generating candidate solutions through random sampling using a minimum number of observations.

While the other techniques use as much data as possible to find the model parameters and then pruning the outliers, RANSAC uses the smallest set of data point possible to estimate the model, thus making it faster and more efficient than the conventional solutions. 

\subsubsection{Finding directional vectors on the object}

In order to find the pose of a planar object, we need to find the three orthonormal vectors on the planar object that describe the object coordinate frame and consequently, the orientation of the object relative to the world coordinate system. We start by estimating the vectors on the planar object that form the basis of the plane, illustrated in figure~\ref{eqn:axis}. Then, we take the cross product of these two vectors to find the third directional vector which is the normal to the object surface. Let's denote the world coordinate system as $XYZ$, and the object coordinate system as $xyz$. We define the axes of the  orientation in relation to a body as: 
\\

\qquad \qquad \qquad \qquad $x \to \text{right}$

\qquad \qquad \qquad \qquad $y \to \text{up}$ 

\qquad \qquad \qquad \qquad $z \to \text{towards the camera}$ 

\begin{figure}[h]       
\centering
    {\includegraphics[width=78pt, height=78pt]{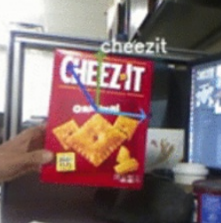}}   
    \hspace{1px}
    {\includegraphics[width=78pt, height=78pt]{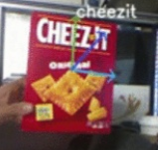}}
    \hspace{1px}
    {\includegraphics[width=78pt, height=78pt]{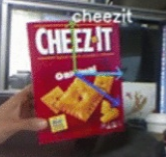}}
    \caption{Computed third directional axis projected onto image plane}
    \label{fig:posevizcam}
\end{figure}

First, we retrieve the locations of the three points $p_c, p_x, p_y$ on the planar object from the reference image using equation (\ref{eqn:axis}) and then locate the corresponding points $p_{c}^{'}, p_{x}^{'}, p_{y}^{'}$ on the image acquired from the Microsoft Kinect sensor. We estimate the locations of these points using the homography matrix $\mathbf{H}$ as shown in equation~\ref{eqn:homography}, \ref{eqn:projection}. Then we find the corresponding 3D locations of $p_{c}^{'}, p_{x}^{'}, p_{y}^{'}$ from the point cloud data also obtained from the Microsoft Kinect sensor. We denote them as vectors $\vec{c}$,$\vec{x}$, and $\vec{y}$. Here, $\vec{c}$ represents the translation vector from the object frame to the world frame and also the position of the object in the world frame. Next, we subtract $\vec{c}$ from $\vec{x}$, $\vec{y}$ which essentially gives us two vectors $\vec{x}$ and $\vec{y}$ centered at the origin of the world frame. We take the cross product of these two vectors $\vec{x}, \vec{y}$ to find the third axis $\vec{z}$. But, depending on the homography matrix the estimated axes $\vec{x}$ and $\vec{y}$ might not be exactly orthogonal, so we take the cross product of $\vec{y}$ and $\vec{z}$ to recalculate the vector $\vec{x}$. Now that we have three orthogonal vectors, we compute the three unit vectors $\hat{i}$, $\hat{j}$, and $\hat{k}$ along the $\vec{x}$, $\vec{y}$, and $\vec{z}$ vectors respectively using equation~\ref{eqn:unitv}. These three orthonormal vectors describe the object frame. These vectors were projected onto the image plane to give a visual confirmation of the methods applied; figure~\ref{fig:posevizcam} shows the orthogonal axes projected onto the object plane.

    

\begin{equation}
\vcenter{\hbox{\begin{minipage}{5cm}
\centering
\includegraphics[width=4cm,height=4cm]{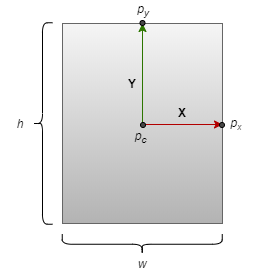}
\captionof{figure}{Axis on the reference plane}
\end{minipage}}}
\begin{aligned}
\left \lbrace \begin{aligned}
p_c &= (w/2, h/2)
\\
p_x &= (w, h/2)
\\
p_y &= (w/2, 0)
\end{aligned} \right . 
\end{aligned}
\label{eqn:axis}
\end{equation}

\begin{align}
\begin{split}
\hat{j} = \frac{\Vec{y}}{|\Vec{y}|} = [j_X \hspace{0.15cm} j_Y \hspace{0.15cm} j_Z]
\\
\hat{k} = \frac{\Vec{x} \times \Vec{y}}{|\Vec{x} \times \Vec{y}|} = [k_X \hspace{0.15cm} k_Y \hspace{0.15cm} k_Z]
\\
\hat{i} = \frac{\Vec{y} \times \Vec{z}}{|\Vec{y} \times \Vec{z}|} = [i_X \hspace{0.15cm} i_Y \hspace{0.15cm} i_Z]
\end{split}
\label{eqn:unitv}
\end{align}

\subsubsection{Planar pose computation}
We compute the pose of the object in terms of the Euler angles. Euler angles are three angles that describe the orientation of a rigid body with respect to a fixed coordinate system. The rotation matrix $\mathbf{R}$ in equation (\ref{eqn:rotR}) rotates X axis to $\hat{i}$, Y axis to $\hat{j}$, and Z axis to $\hat{k}$. 

\begin{align}
 \mathbf{R} = \left [ \begin{matrix} i_X & j_X & k_X \\ i_Y & j_Y & k_Y \\ i_Z & j_Z & k_Z \end{matrix} \right ]
\label{eqn:rotR}   
\end{align}

Euler angles are combinations of the three axis rotations (equation~\ref{eqn:euler-axis}), where $\phi$, $\theta$, and $\psi$ specify the intrinsic rotations around the X, Y, and Z axis respectively.  The combined rotation matrix is a product of three matrices: $\mathbf{R} = \mathbf{R}_z \mathbf{R}_y \mathbf{R}_x$ (equation~\ref{eqn:rotcomb}); the first intrinsic rotation rightmost, last leftmost.

\begin{align}\medmath{
    \left\lbrace \begin{aligned}
\mathbf{R}_x &= \colvec {1 & 0 & 0 \\ 0 & \cos\phi & -\sin\phi \\ 0 & \sin\phi & \cos\phi }  \\
\mathbf{R}_y &= \colvec {\cos\theta & 0 & \sin\theta \\ 0 & 1 & 0 \\ -\sin\theta & 0 & \cos\theta }  \\
\mathbf{R}_z &= \colvec {\cos\psi & -\sin\psi & 0 \\ \sin\psi & \cos\psi & 0 \\ 0 & 0 & 1 }
    \end{aligned} \right .}
    \label{eqn:euler-axis}
\end{align}

\begin{align}
\mathbf{R} = 
    \begin{bmatrix*}
        c\theta c\psi
        &  s\phi s\theta c\psi - c\phi s\psi
        &  c\phi s\theta c\psi + s\phi s\psi
        \\ c\theta s\psi
        &  s\phi s\theta s\psi + c\phi c\psi
        &  c\phi s\theta s\psi - s\phi c\psi
        \\ -s\theta
        &  s\phi c\theta
        &  c\phi c\theta
    \end{bmatrix*}
\label{eqn:rotcomb}
\end{align}

In equation~\ref{eqn:rotcomb}, $c$ and $s$ represents $\cos$ and $\sin$ respectively.

Solving for $\phi, \theta$, and $\psi$ from (\ref{eqn:rotR}) and (\ref{eqn:rotcomb}), we get,

\begin{align}\medmath{
    \left\lbrace \begin{aligned}
    \phi &= \tan^{-1}\left(\frac{j_Z}{k_Z}\right) \\
    \theta &= \tan^{-1}\left(\frac{-i_Z}{\sqrt{1-i_Z^2}}\right) = \sin^{-1}\left(-i_Z\right) \\
    \psi &= \tan^{-1}\left(\frac{i_Y}{i_X}\right)
    \end{aligned} \right .}
    \label{eqn:eulerangles}
\end{align}

\begin{figure*}
\centering
\subfloat[]{\includegraphics[width = 135pt, height=80pt]{{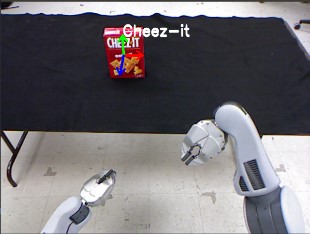}}}\hspace{10px}
\subfloat[]{\includegraphics[width = 135pt, height=80pt]{{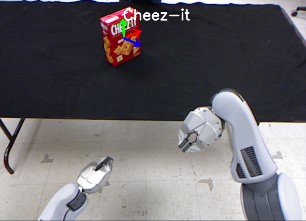}} }\hspace{10px}
\subfloat[]{\includegraphics[width = 135pt, height=80pt]{{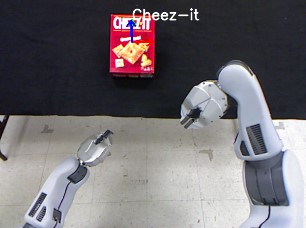}}}

\vspace{0pt}
\subfloat[]{\includegraphics[width = 135pt, height=80pt]{{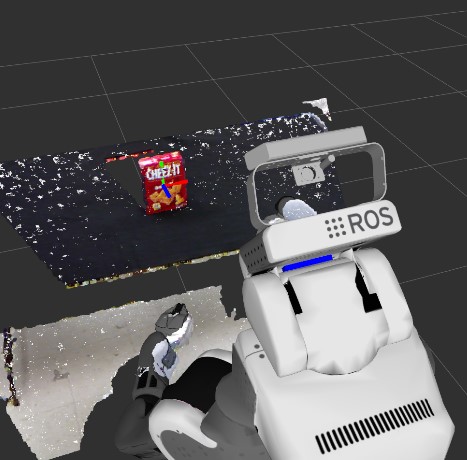}}} \hspace{10px}
\subfloat[]{\includegraphics[width = 135pt, height=80pt]{{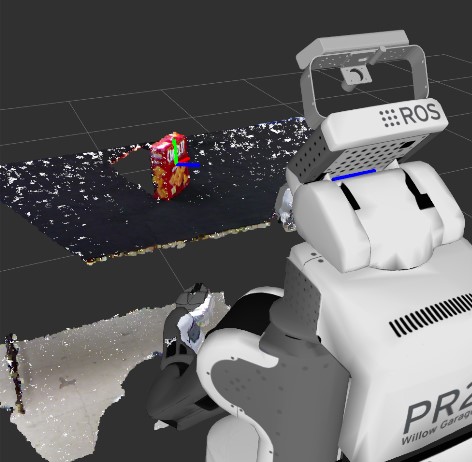}}} \hspace{10px}
\subfloat[]{\includegraphics[width = 135pt, height=80pt]{{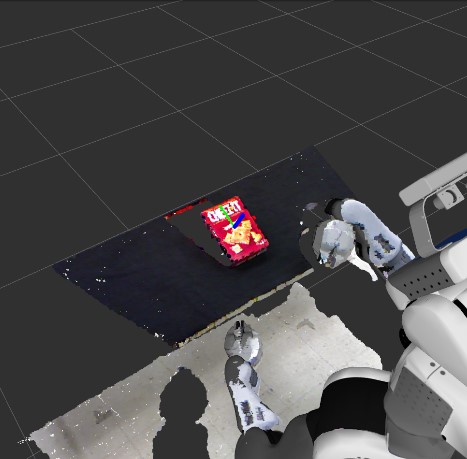}}}

\vspace{-5pt}
   \caption{(a),(b),(c) are recovered poses from robot's camera and (d),(e),(f) are corresponding poses visualized in RViz}
\label{fig:poseviz}
\end{figure*}

\subsection{Training Grasps for Humanoid Robots}
\begin{figure}
    \centering
    \includegraphics[width=0.65\linewidth]{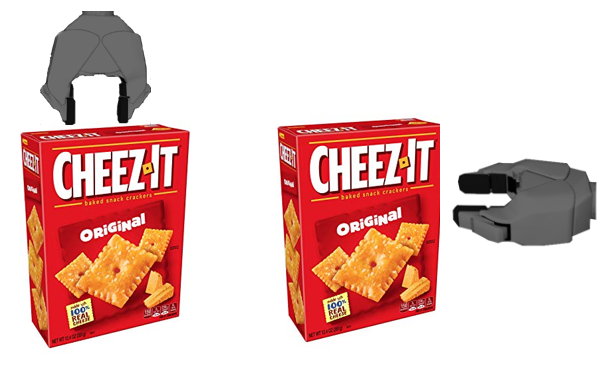}
    \caption{Pre-training canonical grasp}
    \label{fig:can_grasp}
\end{figure}

To ensure that the robot can grasp objects in an adaptive manner, we pre-train the robot to perform a set of canonical grasps. We place the object and the robot's gripper close to each other and record the relative pose. This essentially gives us the pose of the gripper with respect to the object. Figure~\ref{fig:can_grasp} illustrates the training process in which the robot's gripper and a cracker box have been placed in close proximity and the relative poses have been recorded for grasping the objects from the side. 

\begin{equation}
    \textbf{T}_{s}^{d}
    = \begin{bmatrix} \textbf{R}_{s}^{d} & P_{s}^{d} \\ 0 & 1 \end{bmatrix}
    =\begin{bmatrix} r_{11} & r_{12} & r_{13} & X_t \\
                r_{21} & r_{22} & r_{23} & Y_t \\
                r_{31} & r_{32} & r_{33} & Z_t \\
                0 & 0 & 0 & 1 \end{bmatrix} 
\label{eqn:transmat}
\end{equation}

Equation~\ref{eqn:transmat} outlines the structure of a transformation matrix $\textbf{T}_{s}^{d}$ that describes the rotation and translation of frame  $d$ with respect to frame $s$; $\textbf{R}_{s}^{d}$ represents the rotation matrix similar to equation~\ref{eqn:rotcomb} and $P_{s}^{d}=[X_{t},Y_{t},Z_{t}]^{T}$ is the translation matrix which is the 3D location of the origin of frame $d$ in frame $s$.

During the training phase, we first formulate the transformation matrix $\textbf{T}_{b}^{o}$ using the rotation matrix and the object location. We take the inverse of $\textbf{T}_{b}^{o}$ which gives us the transformation matrix $\textbf{T}_{o}^{b}$. We then use the equation~\ref{eqn:graspmat} to record the transformation $\mathbf{T}_{o}^{g}$ of the robot's wrist relative to the object.

\begin{equation} \label{eqn:graspmat}
T_{o}^{g} = T_{o}^{b} \times T_{b}^{g} \  \text{where} \ T_{o}^{b} = (T_{b}^{o})^{-1}
\end{equation}
\par

In the equation~\ref{eqn:graspmat}, $b$ refers to the robot's base, $o$ refers to the object, and $g$ refers to the wrist of the robot to which the gripper is attached. Once we record the matrix, we get a new pose of the object from the vision in the testing phase and generate the final matrix using the equation~\ref{eqn:fingrasp} that has the new position and orientation of the robot's wrist in matrix form .
\begin{equation} \label{eqn:fingrasp}
T_{b}^{g} = T_{b}^{o} \times T_{o}^{g}
\end{equation}
\par

We then extract the rotational angles $\gamma$, $\beta$, $\alpha$~(roll, pitch, yaw)  of the grasp pose from matrix $\mathbf{T}_{b}^{g}$ using equation~\ref{eqn:grippereulerangles}

\begin{align}\medmath{
    \left\lbrace \begin{aligned}
    \gamma=tan^{-1}(r_{32}/r_{33}) \\
    \beta=tan^{-1}\frac{-r_{31}}{\sqrt {{r_{32}}^2+{r_{33}}^2}}\\
    \alpha=tan^{-1}(r_{21}/r_{11})
    \end{aligned} \right .}
    \label{eqn:grippereulerangles}
\end{align}

\section{Evaluation}
The proposed object recognition and pose estimation algorithm was implemented on an Ubuntu 14.04 platform equipped with 3.0 GHz Intel R Core(TM) i5-7400 CPU and 8GB system memory. The RGB-D camera used in the experiments was a Microsoft Kinect sensor v1. We evaluated the proposed algorithm by comparing the accuracy of object recognition, pose estimation, and execution time of four different feature descriptors. We also validated the effectiveness of our approach for adaptive grasping by conducting experiments with the PR2 robot.

\subsection{Object detection and pose estimation}

\begin{figure}[h]
\begin{center}
\graphicspath{ {./images/} }
\includegraphics[width=5cm]{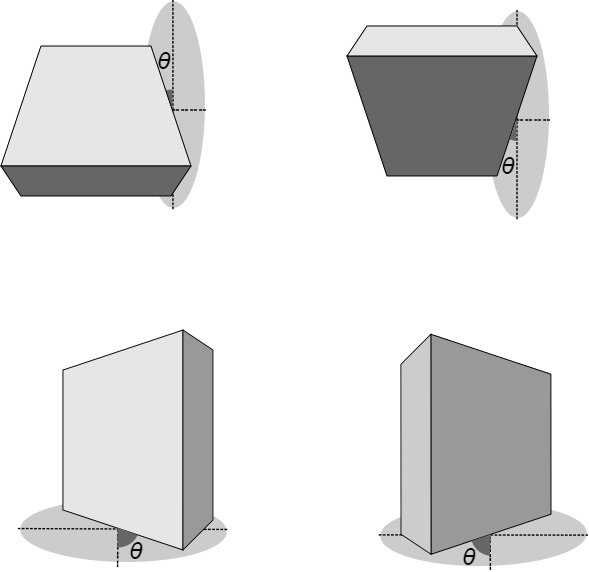}
\end{center}
  \caption{Out of plane rotation}
\label{fig:theta}
\end{figure}

Without enough observable features, the system would fail to find good matches that are required for accurate homography estimation. Consequently, our object detection and pose estimation approach has a constraint on the out-of-plane rotation $\theta$, illustrated in figure~\ref{fig:theta}. In other words, if the out-of-plane rotation of the object is more than $\theta$, the system would not be able to recognize the object. Fast execution is also a crucial aspect to facilitate multiple object detection and pose estimation for real-time applications. We experimented with four different descriptors on several planar objects and the comparative result is shown in table~\ref{tbl:comparisondescriptor}. The execution time was measured for the object detection and pose estimation step. AKAZE and BRISK had much lower processing time for detection and pose estimation, thus would have a better frame rate, but SIFT and SURF had larger out-of-plane rotational freedom.

\begin{table}[h!]
\centering
    \caption{Comparison of feature descriptors}
    \begin{tabular}{l|c|s}
      \toprule 
      \textbf{Descriptor} & \begin{tabular}[x]{@{}c@{}}\textbf{Maximum out of}\\ \textbf{plane rotation} (degree)\end{tabular} & \begin{tabular}[x]{@{}c@{}}\textbf{Execution time}\\ (second)\end{tabular}\\
      \midrule 
      
      SIFT & $48^{\circ}\pm2^{\circ}$ & \text{~~~~~~~0.21s}\\ \hline
      SURF & $37^{\circ}\pm2^{\circ}$ & \text{~~~~~~~0.27s}\\ \hline
      AKAZE & $18^{\circ}\pm1^{\circ}$ & \text{~~~~~~~0.05s}\\ \hline
      BRISK & $22^{\circ}\pm2^{\circ}$ & \text{~~~~~~~0.06s}\\
      \bottomrule 
    \end{tabular}
    \label{tbl:comparisondescriptor}
\end{table}

\begin{figure}[h]
\begin{center}
\graphicspath{ {./images/} }
\includegraphics[width=\linewidth]{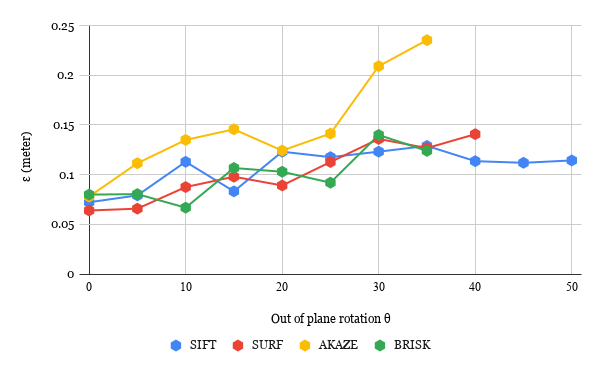}
\end{center}
  \caption{Out of plane rotation vs $\epsilon$}
\label{fig:epsilon}
\end{figure}

We also compared the \textit{RMS} difference $\epsilon$~(equation~\ref{eqn:epsilon}) of re-calculated $\vec{x}$ to original $\vec{x}$ ($\vec{x}^{'}$ in the equation) for increasing out-of-plane rotation of the planar objects to assess the homography estimation. Ideally, the two estimated vectors $\vec{x}$ and $\vec{y}$, which describe the basis of the plane of the planar object, should be orthogonal to each other, but often they are not. So, the values of $\epsilon$ in figure~\ref{fig:epsilon} give us an indication of the average error in homography estimation for different out-of-plane rotations. In figure~\ref{fig:epsilon}, we can see AKAZE has much higher $\epsilon$ values while the rest remained within a close range. This tells us AKAZE results in a much larger error in estimating the homography than the other methods. 

We chose SIFT and SURF to evaluate how the execution time for detection scales up while increasing the number of objects. From table~\ref{tbl:multobjcompdesc}, which shows the mean processing time for object detection, we can see that SURF had a detection time around 50\% more than SIFT in all the cases. This outcome coupled with the previous results prompted us to select SIFT for the subsequent experiments.

\begin{figure}[h]       
\centering
    {\includegraphics[width=78pt, height=78pt]{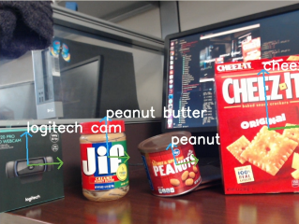}}   
    \hspace{1px}
    {\includegraphics[width=78pt, height=78pt]{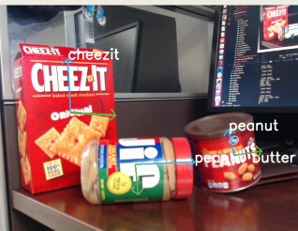}}
    \hspace{1px}
    {\includegraphics[width=78pt, height=78pt]{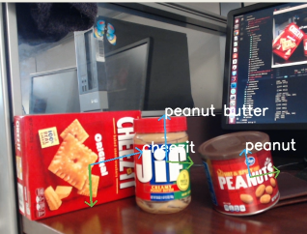}}
    \caption{\centering Multiple object detection with estimated planar vectors}
    \label{fig:multobjdet}
\end{figure}

The system was capable of detecting multiple objects in real-time and at the same time could estimate their corresponding poses. Figure~\ref{fig:multobjdet} shows detected objects with estimated directional planar vectors. We can also observe that the system was robust to in-plane rotation and partial occlusion.

\begin{table}[h]
\centering
\caption{\centering Execution time of SIFT and SURF for multiple object detection}
\begin{tabular}{|c|c|c|}
\hline
\multirow{2}{*}{\begin{tabular}[c]{@{}c@{}}Number of\\ Objects\end{tabular}} & \multicolumn{2}{c|}{\begin{tabular}[c]{@{}c@{}}Detection time\\ (second)\end{tabular}} \\ \cline{2-3} 
                                                                             & SIFT                                       & SURF                                       \\ \hline
1                                                                            & 0.06s                                       & 0.09s                                       \\ \hline
2                                                                            & 0.11s                                       & 0.17s                                       \\ \hline
3                                                                            & 0.17s                                       & 0.26s                                       \\ \hline
4                                                                            & 0.22s                                       & 0.35s                                       \\ \hline
5                                                                            & 0.28s                                       & 0.4s5                                       \\ \hline
6                                                                            & 0.34s                                       & 0.54s                                       \\ \hline
\end{tabular}
\label{tbl:multobjcompdesc}
\end{table}

\begin{figure}[h]       
\centering
    {\includegraphics[width=100pt, height=100pt]{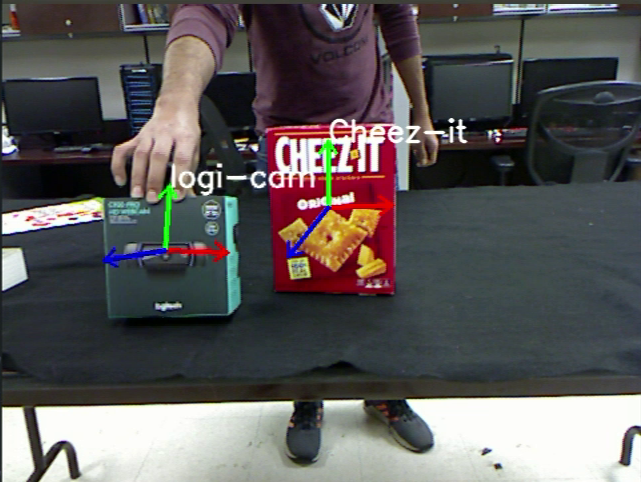}}   
    \hspace{1px}
    {\includegraphics[width=100pt, height=100pt]{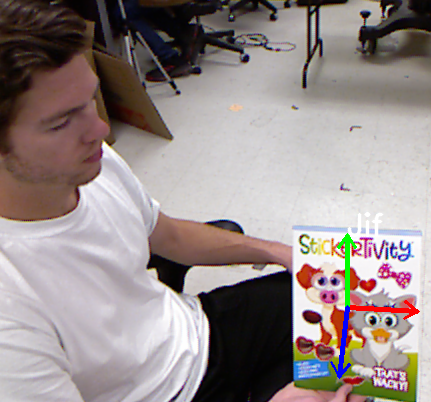}}
    \caption{(a) Pose estimation of multiple objects (b) Estimated pose of an object held by a human}
    \label{fig:pose_check}
\end{figure}

We used RViz~\cite{RViz}, a 3D visualizer for the Robot Operating System (ROS)~\cite{ros}, to validate the pose estimation. The calculated directional axes were projected onto the image and the estimated poses were visualized in RViz. As shown in figure~\ref{fig:poseviz}, we qualitatively verified the accuracy of the detection and the estimated pose by comparing the two outputs. We can see that both the outputs render similar results. We conducted experiments with multiple objects and human held objects as well. Figure~\ref{fig:pose_check} illustrates the simultaneous detection and pose estimation of two different boxes and an object held by a human, respectively.

\begin{equation}
    \epsilon = \frac{1}{N}\sum_{i=1}^{N}||\vec{x_i}^{'}-\vec{x_i}||, \text{\fontsize{8}{8}\selectfont where N is the number of frames}
    \label{eqn:epsilon}
\end{equation}

\subsection{Adaptive grasping}
We assessed our approach for adaptive grasping keeping two different aspects of the robotic application in mind; robotic tasks that require 1) interacting with a static environment, and 2) interacting with humans.

\begin{figure}
    \centering
    \captionsetup[subfigure]{labelformat=empty}
    
    \subfloat[]{\includegraphics[width=0.3\linewidth, height=0.3\linewidth]{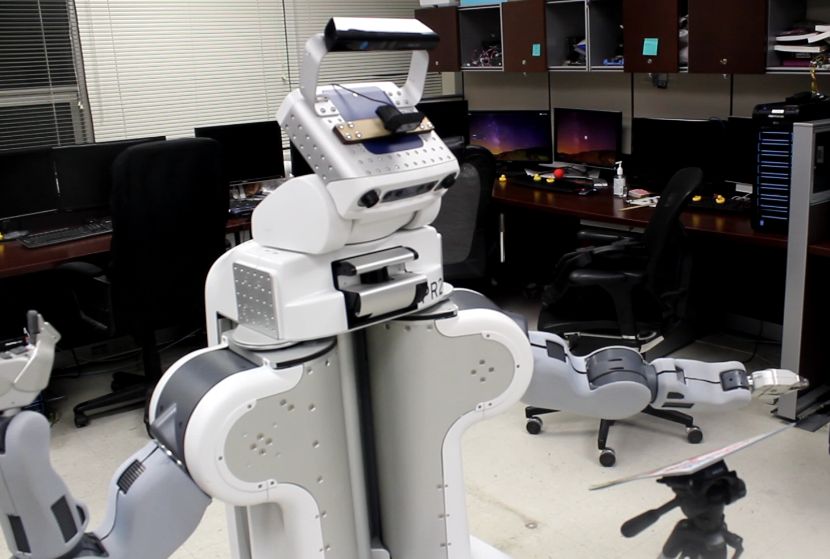}}
    \vspace{5px}
    \subfloat[]{\includegraphics[width=0.3\linewidth, height=0.3\linewidth]{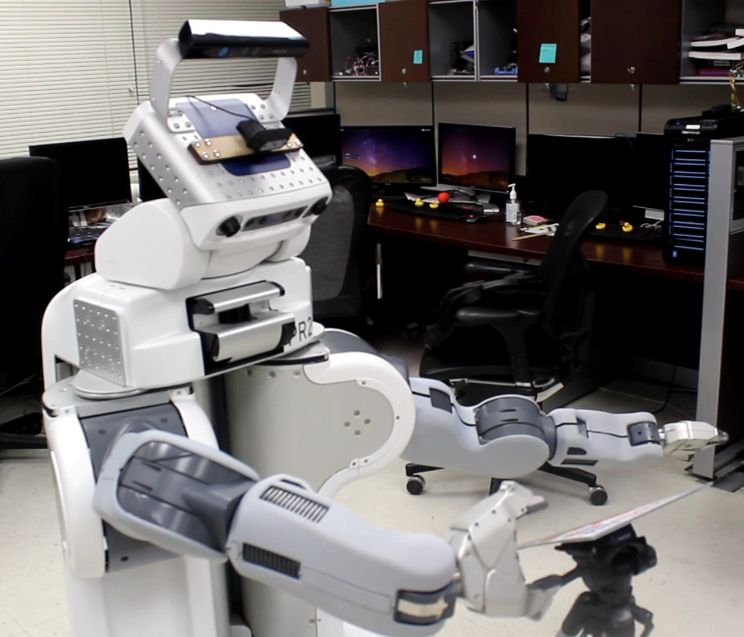}}
    \vspace{5px}
    \subfloat[]{\includegraphics[width=0.3\linewidth, height=0.3\linewidth]{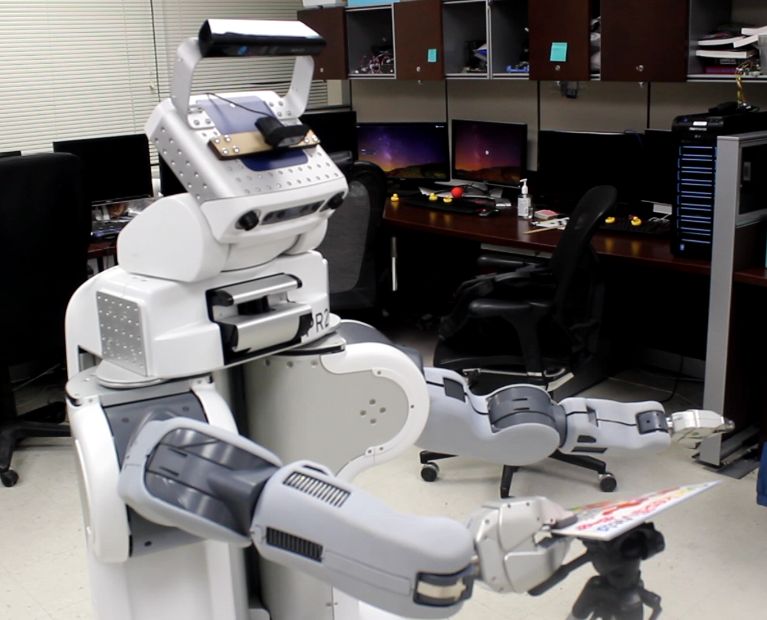}}\\[-5ex]
    
    \subfloat[]{\includegraphics[width=0.3\linewidth, height=0.3\linewidth]{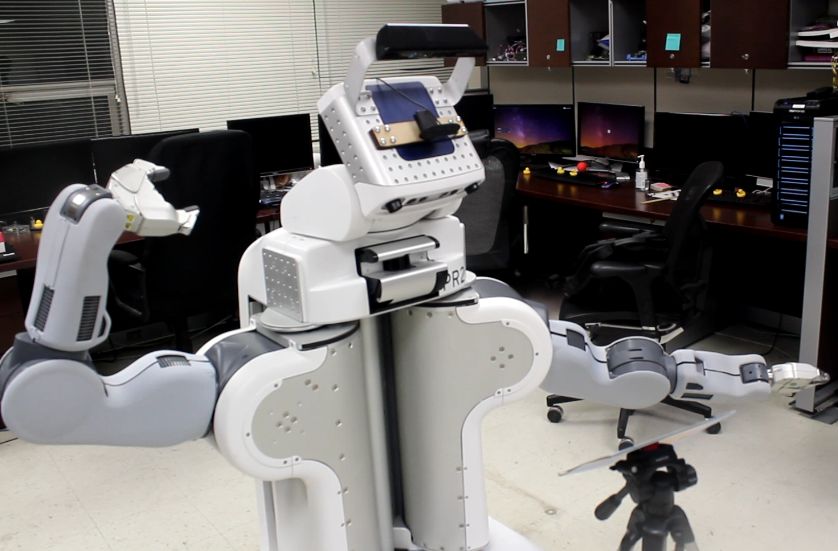}}
    \vspace{5px}
    \subfloat[]{\includegraphics[width=0.3\linewidth, height=0.3\linewidth]{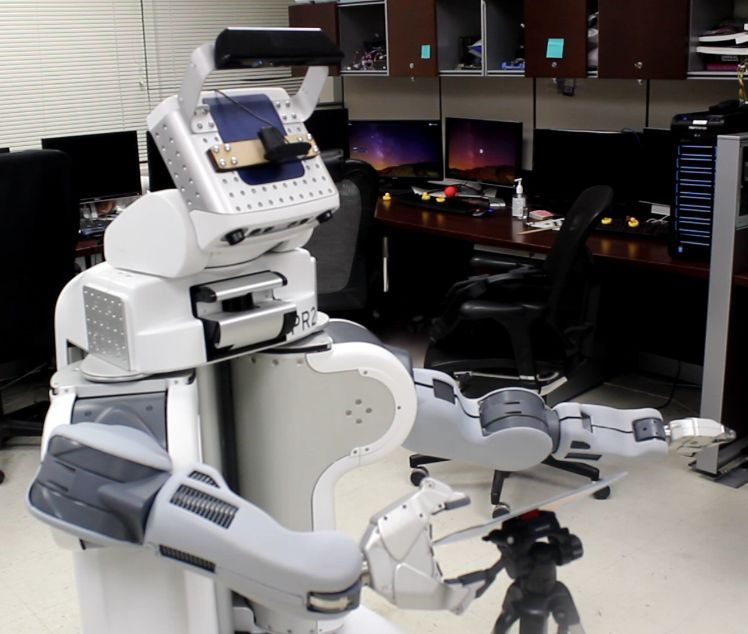}}
    \vspace{5px}
    \subfloat[]{\includegraphics[width=0.3\linewidth, height=0.3\linewidth]{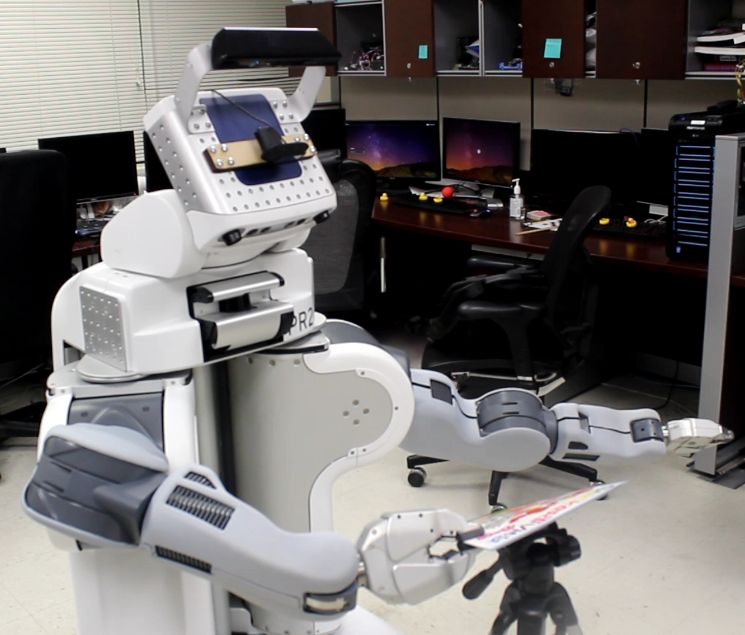}}\\[-5ex]
    
    \subfloat[]{\includegraphics[width=0.3\linewidth, height=0.3\linewidth]{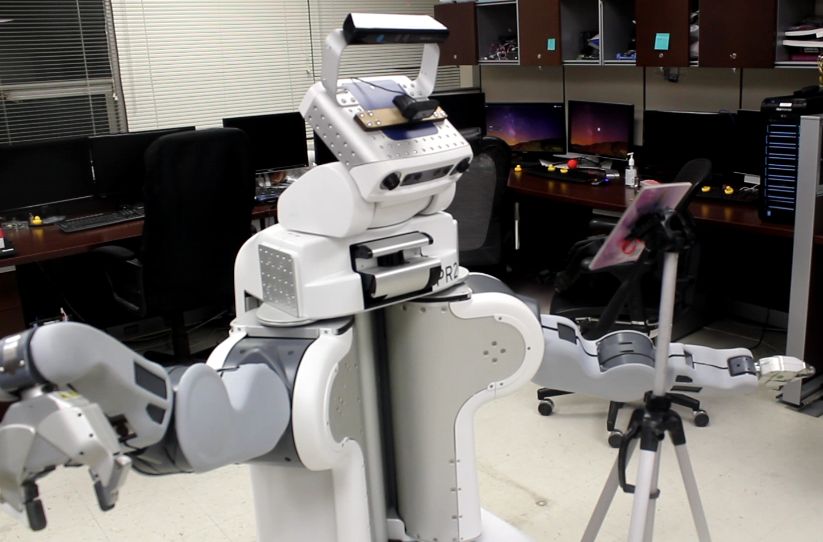}}
    \vspace{5px}
    \subfloat[]{\includegraphics[width=0.3\linewidth, height=0.3\linewidth]{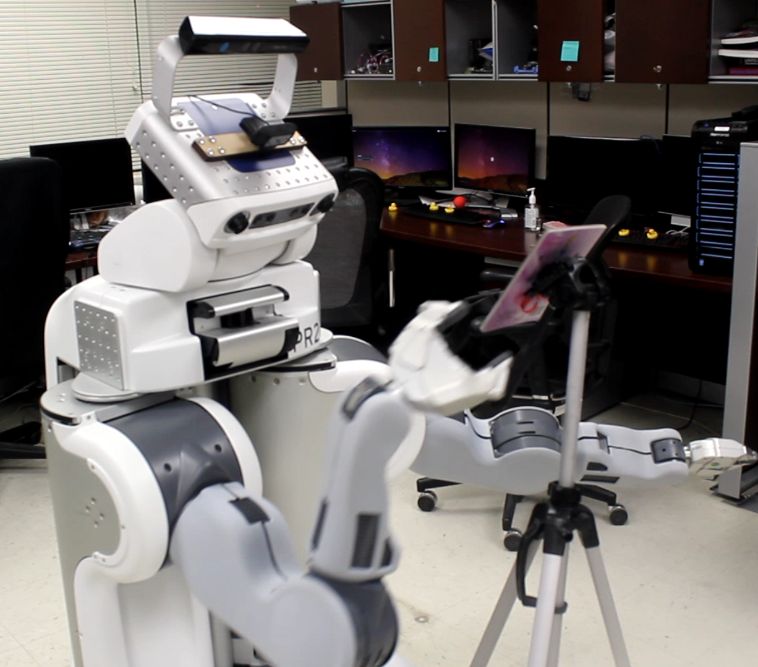}}
    \vspace{5px}
    \subfloat[]{\includegraphics[width=0.3\linewidth, height=0.3\linewidth]{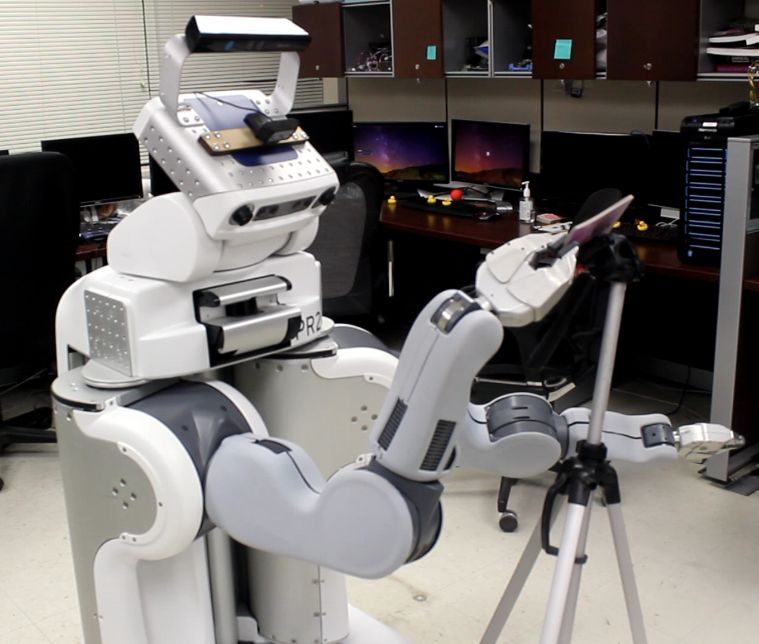}}
    
    \caption{Robot grasping an object from a tripod. Left: initial position of the robot's gripper, middle: gripper adapting to the object's pose, right: grasping of the object.}
    \label{fig:tripod_results}
\end{figure}

\begin{figure}
    \centering
    \captionsetup[subfigure]{labelformat=empty}
    
    \subfloat[]{\includegraphics[width=0.30\linewidth, height=0.35\linewidth]{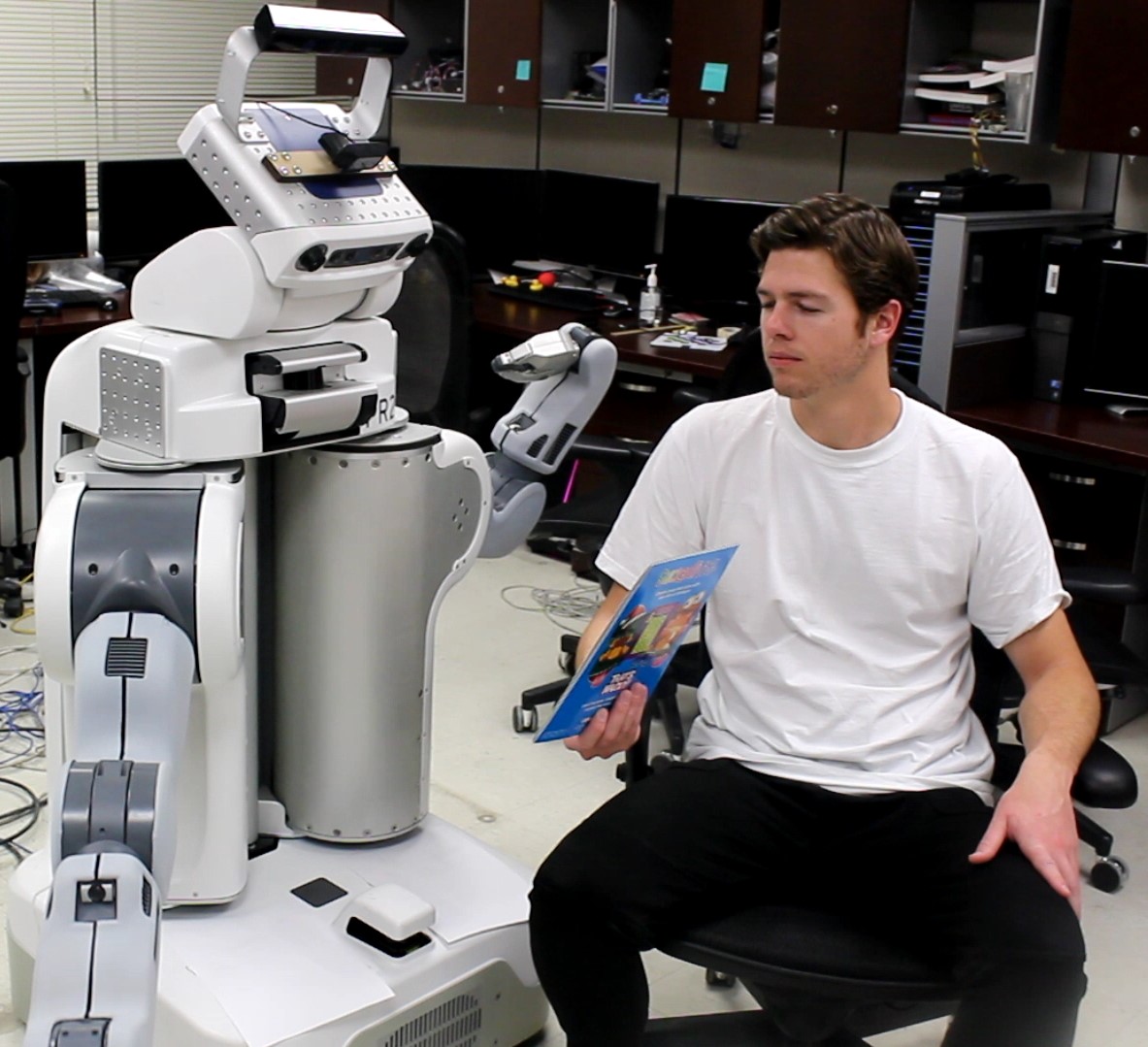}}
    \vspace{5px}
    \subfloat[]{\includegraphics[width=0.30\linewidth, height=0.35\linewidth]{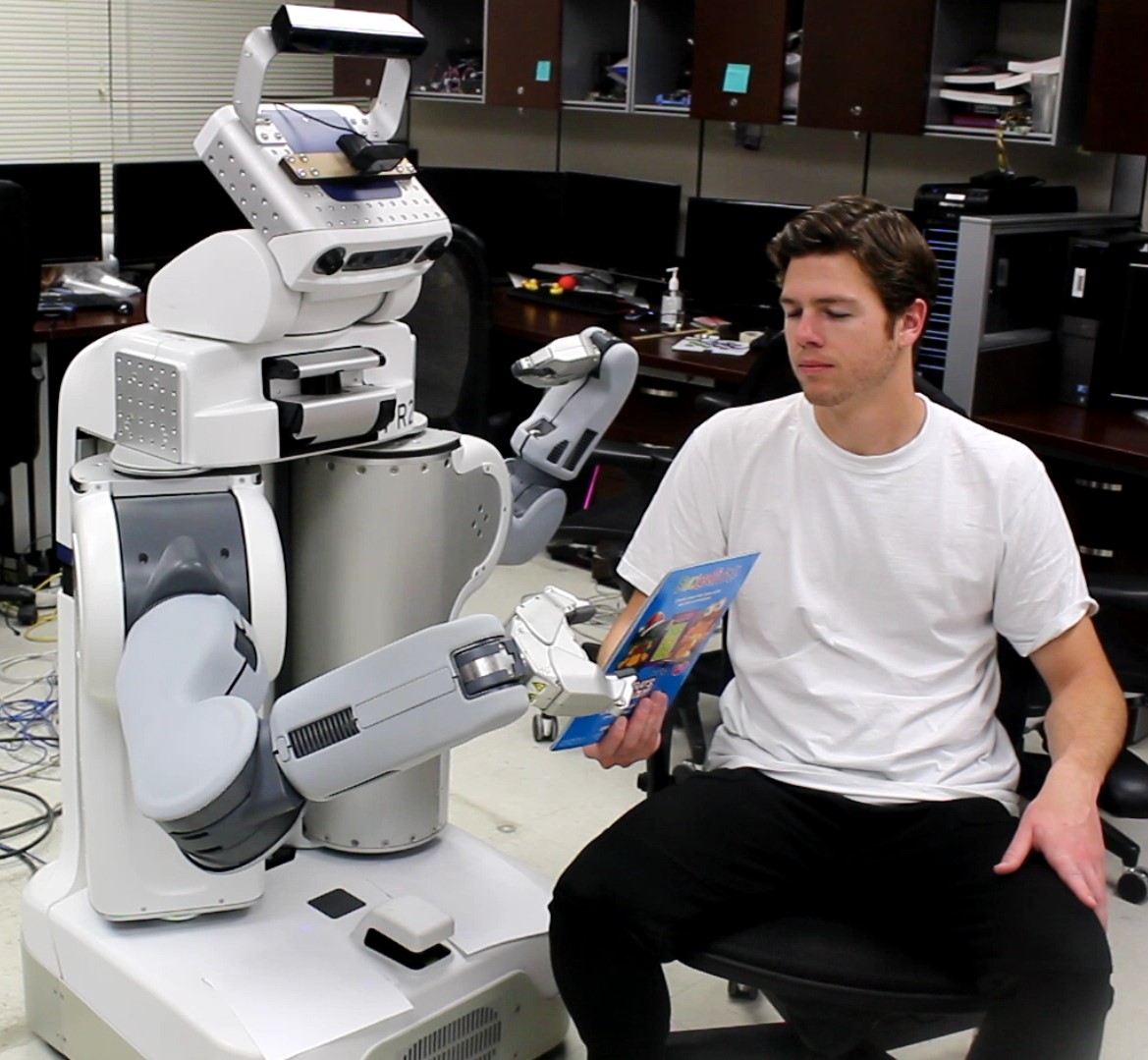}}
    \vspace{5px}
    \subfloat[]{\includegraphics[width=0.30\linewidth, height=0.35\linewidth]{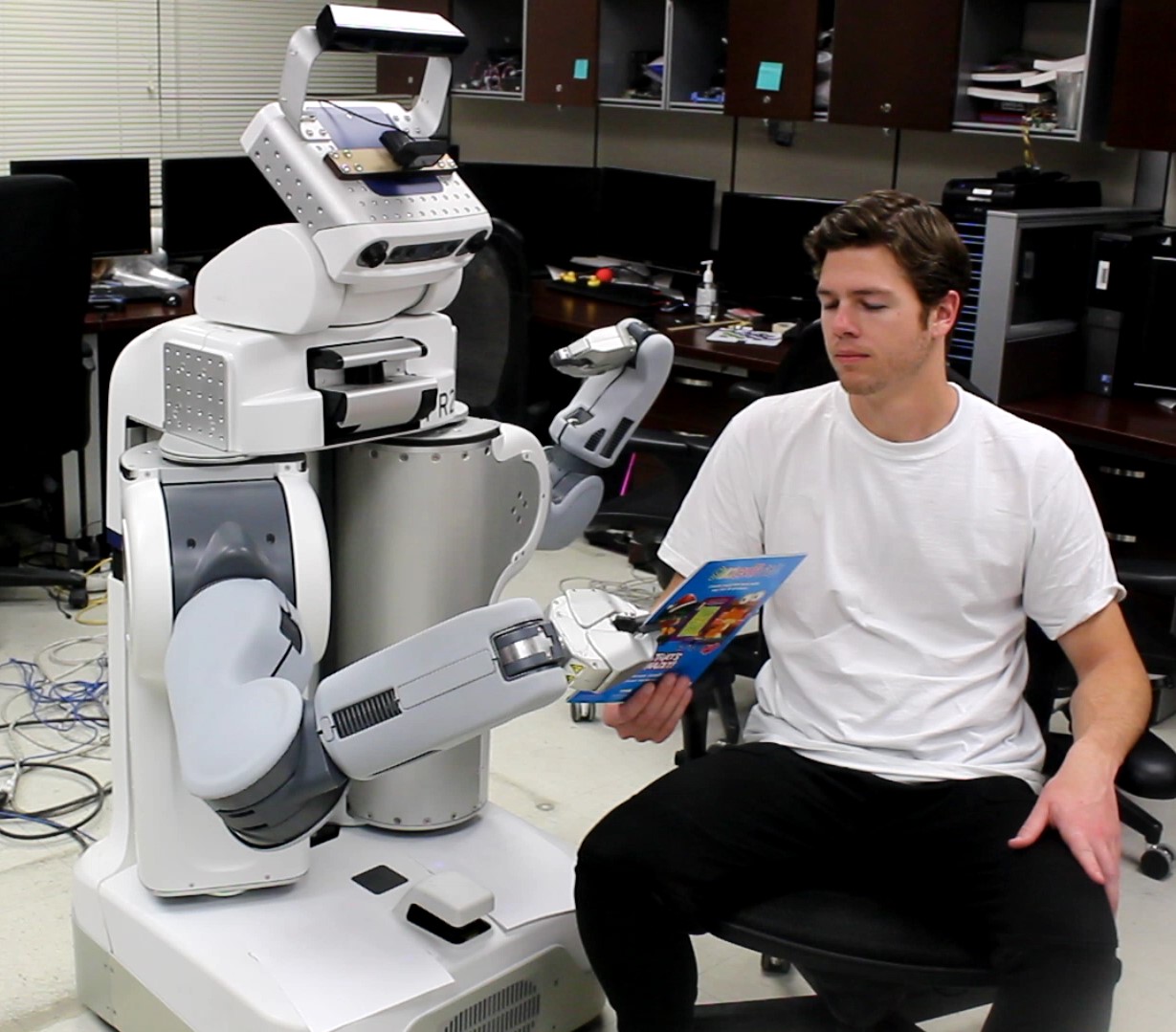}}\\[-5ex]
    
    \subfloat[]{\includegraphics[width=0.30\linewidth, height=0.35\linewidth]{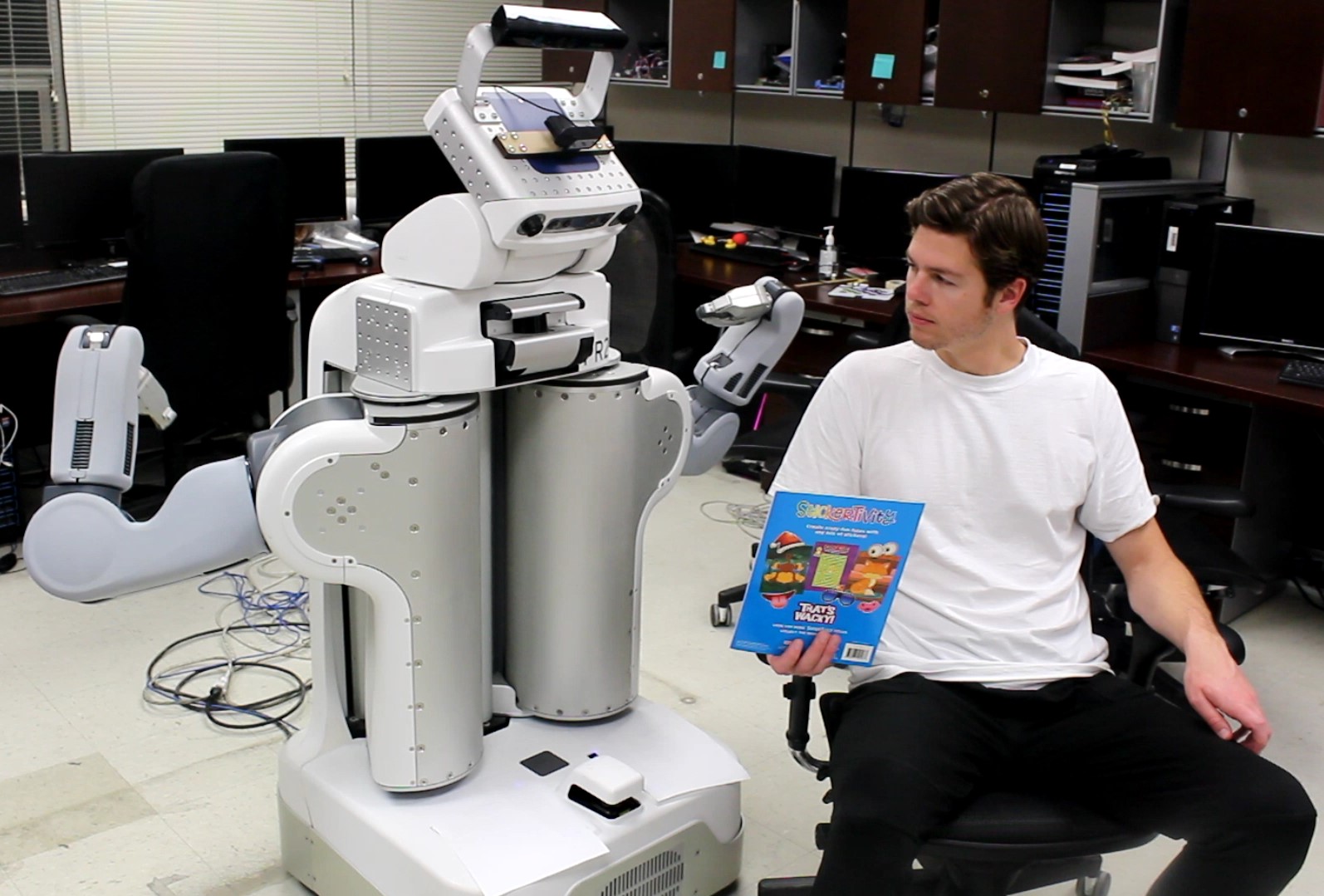}}
    \vspace{5px}
    \subfloat[]{\includegraphics[width=0.30\linewidth, height=0.35\linewidth]{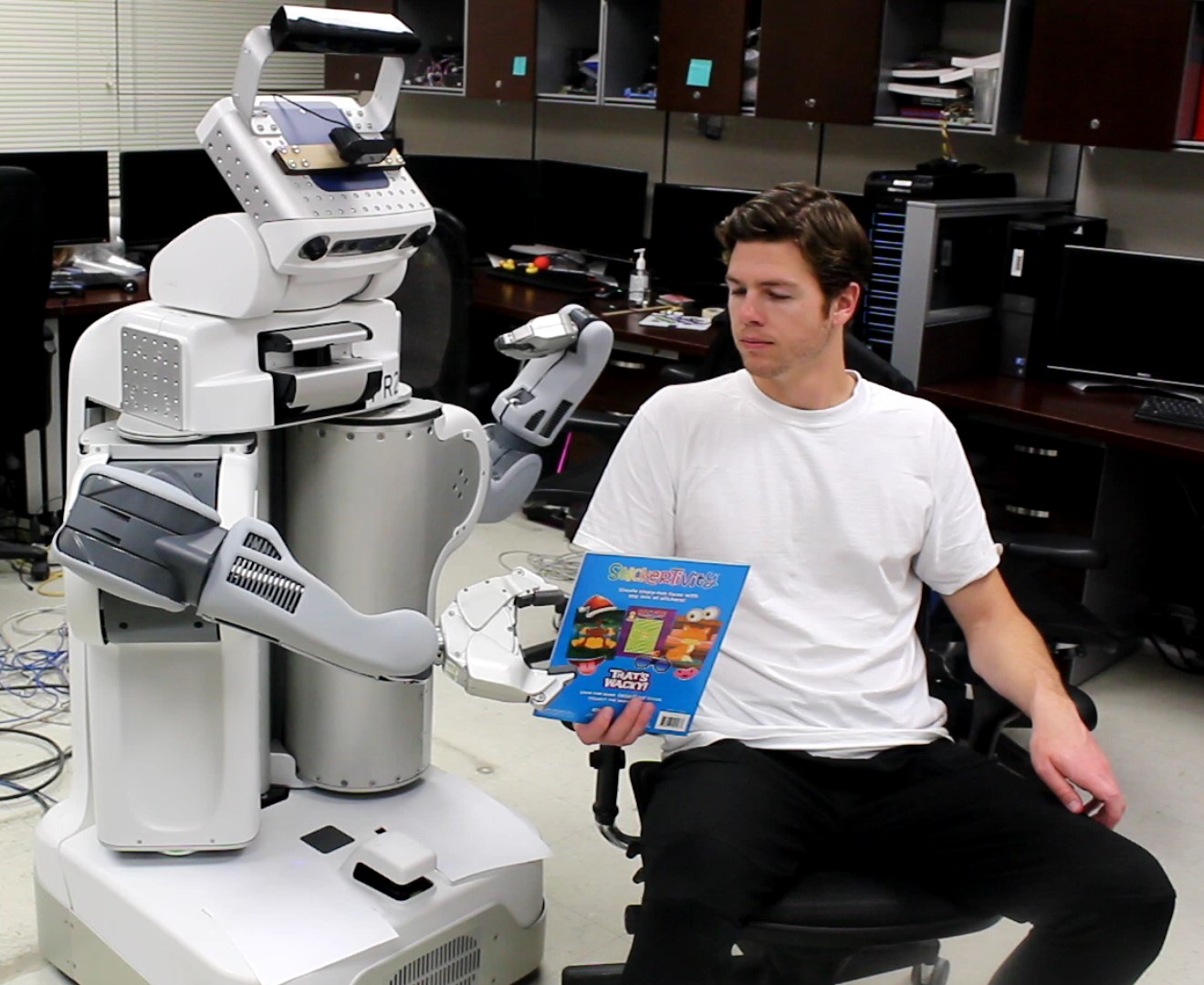}}
    \vspace{5px}
    \subfloat[]{\includegraphics[width=0.30\linewidth, height=0.35\linewidth]{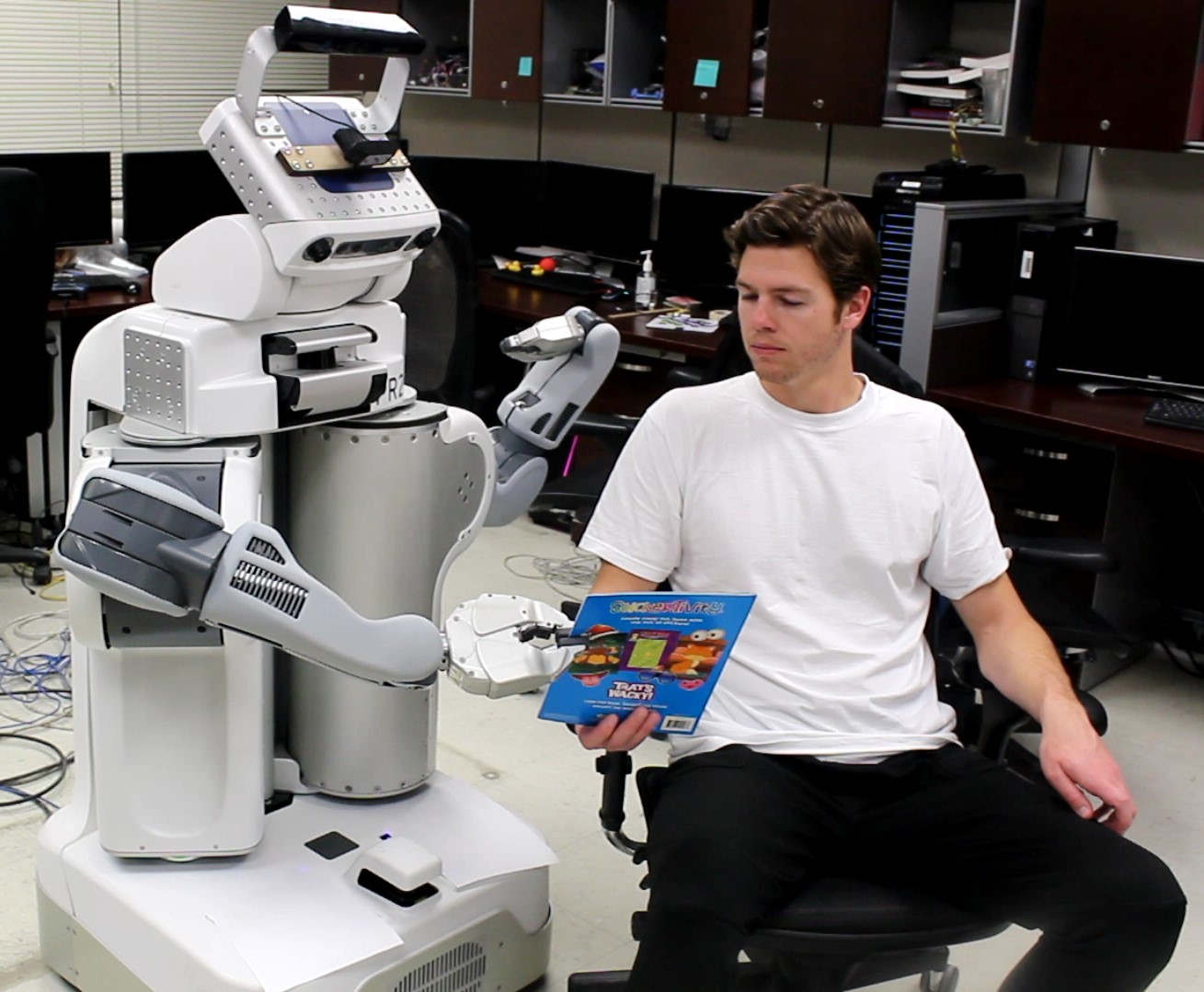}}\\[-5ex]
    
    \subfloat[]{\includegraphics[width=0.30\linewidth, height=0.35\linewidth]{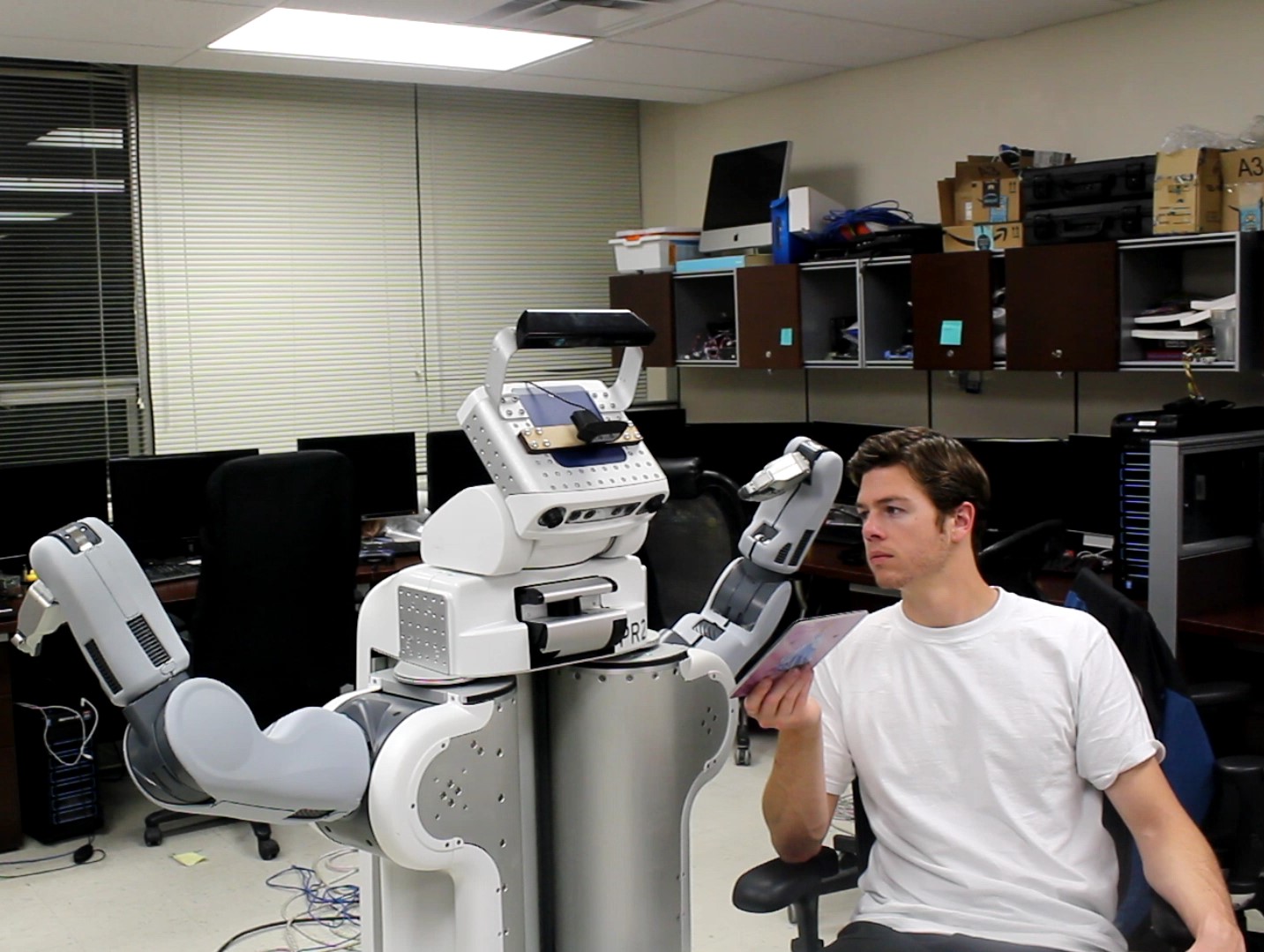}}
    \vspace{5px}
    \subfloat[]{\includegraphics[width=0.30\linewidth, height=0.35\linewidth]{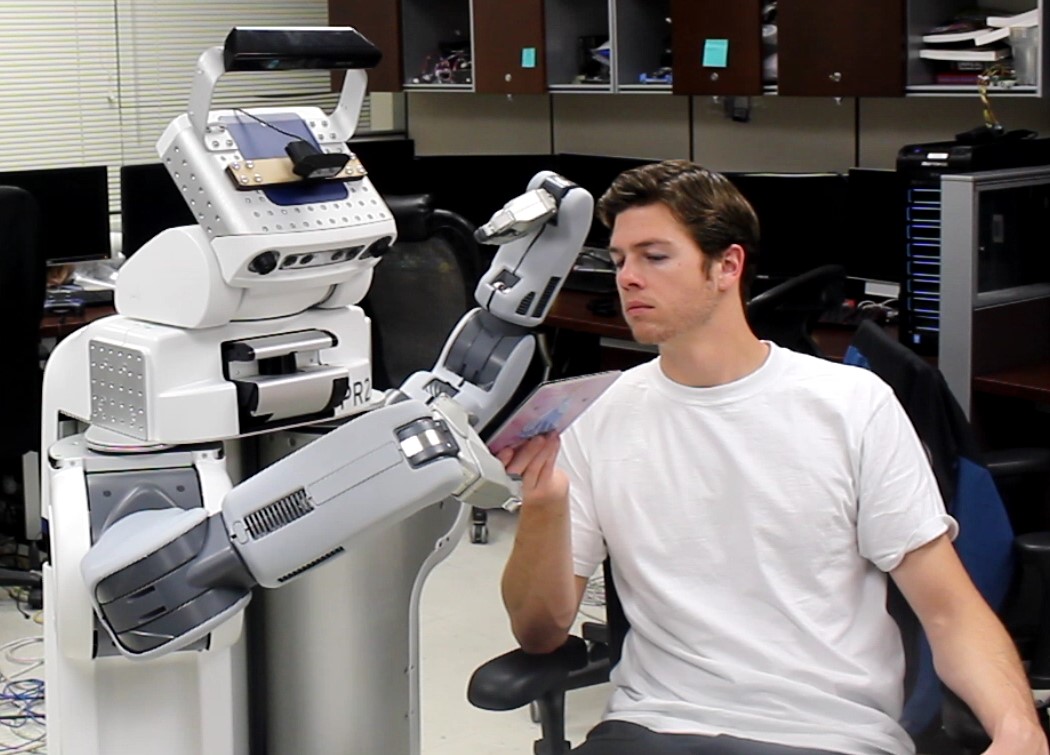}}
    \vspace{5px}
    \subfloat[]{\includegraphics[width=0.30\linewidth, height=0.35\linewidth]{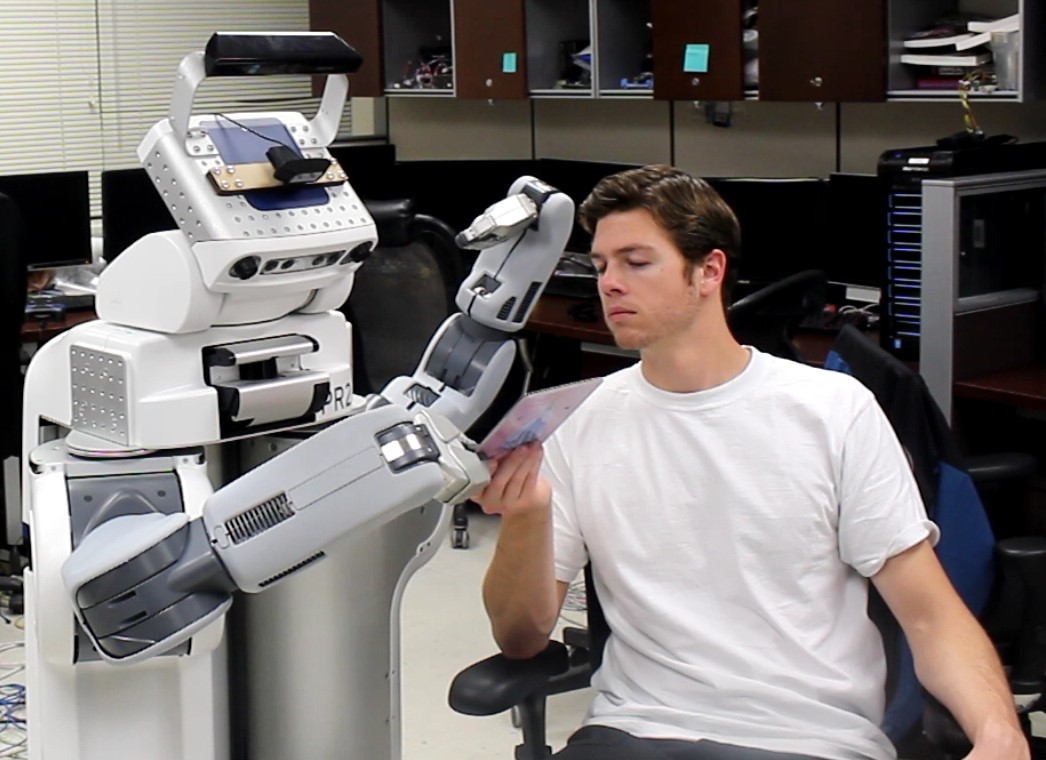}}
    
    \caption{Robot grasping an object held by a human. Left: initial position of the robot's gripper, middle: gripper adapting to the object's pose, right: grasping of the object.}
    \label{fig:human_results}
\end{figure}

We first tested our system for static objects where the object was attached to a tripod. Next, we set up experiments where the object was held by a human. We used a sticker book and a cartoon book and evaluated our system on a comprehensive set of poses. In almost all the experiments, the robot successfully grasped the object in a manner consistent with its training. There were some poses that were not reachable by the robot - for instance, when the object was pointing inward along the X-axis in the robot reference frame, it was not possible for the end-effector to make a top grasp. Figure~\ref{fig:tripod_results} and \ref{fig:human_results} show the successful grasping of the robot for both types of experiments.

\section{Conclusion and Future Work}

This work presents an approach that enables humanoid robots to grasp objects using planar pose estimation based on RGB image and depth data. We examined the performance of four feature-detector-descriptors for object recognition and found SIFT to be the best solution. We used FLANN's K-d Tree Nearest Neighbor implementation, and Bruteforce Hamming to find the keypoint matches and employed RANSAC to estimate the homography. The homography matrix was used to approximate the three orthonormal directional vectors on the planar object using perspective transformation. The pose of the planar object was estimated from the three directional vectors. The system was able to detect multiple objects and estimate the pose of the objects in real-time. We also conducted experiments with the humanoid PR2 robot to show the practical applicability of the framework where the robot grasped objects by adapting to a range of different poses.

In the future, we plan to add GPU acceleration for the proposed algorithm that would further improve the overall computational efficiency of the system. We would like to extend the algorithm to automatically prioritize certain objects and limit the number of objects needed for detection based on different scheduled tasks. Finally, we would like to incorporate transferring grasp configuration for familiar objects and explore other feature matching technique e.g. multi probe LSH, hierarchical k-means tree, etc.

 \bibliographystyle{unsrt2}
 
 \footnotesize{
 \bibliography{egbib}
 }



\end{document}